\documentclass[conference]{IEEEtran}
\IEEEoverridecommandlockouts
\usepackage{cite}
\usepackage{amsmath,amssymb,amsfonts}
\usepackage{algorithmic}
\usepackage{graphicx}
\usepackage{textcomp}
\usepackage[dvipsnames]{xcolor}
\usepackage{float}
\usepackage{subfig}
\usepackage{caption}
\usepackage{multirow}

\newcommand{\ian}{\textcolor{black}}
\newcommand{\iann}{\textcolor{black}}

\DeclareMathOperator{\EX}{\mathbb{E}}

\begin{document}

\title{Human-Like Navigation Behavior: \\ A Statistical Evaluation Framework}

\author{\IEEEauthorblockN{Ian Colbert}
\IEEEauthorblockA{\textit{Advanced Micro Devices, Inc.} \\
San Diego, United States \\
ian.colbert@amd.com}
\and
\IEEEauthorblockN{Mehdi Saeedi}
\IEEEauthorblockA{\textit{Advanced Micro Devices, Inc.} \\
Markham, Canada \\
mehdi.saeedi@amd.com}}

\maketitle

\begin{abstract}
Recent advancements in deep reinforcement learning have brought forth an impressive display of highly skilled artificial agents capable of complex intelligent behavior.
In video games, these artificial agents are increasingly deployed as non-playable characters (NPCs) designed to enhance the experience of human players.
However, while it has been shown that the convincing human-like behavior of NPCs leads to increased engagement in video games, the believability of an artificial agent's behavior is \ian{most} often measured solely by its proficiency at a given task.
Recent work has hinted that proficiency alone is not sufficient to discern human-like behavior.
Motivated by this, we build a non-parametric two-sample hypothesis test designed to compare the behaviors of artificial agents to those of human players.
We show that the resulting $p$-value not only aligns with anonymous human judgment of human-like behavior, but also that it can be used as a measure of similarity.
\end{abstract}


\section{Introduction}
\label{sec:intro}

Of the many research topics within the field of artificial intelligence (AI) for gaming, the development of believable non-playable characters (NPCs) has received particular attention due to the increasingly impressive successes in training highly skilled artificial agents to play complex games \ian{using deep reinforcement learning}~\cite{alonso2020deep, pearce2021counter, team2021open}.
However, while the believability of an artificial agent is often contextually measured by its proficiency in accomplishing a specifically designed goal, the \textit{behaviors} that an agent exhibits in the process is equally as important.
In fact, it has been shown that developing artificial agents to exhibit human-like behavior leads to increased engagement in games~\cite{soni2008bots}.
\ian{Yet, the focus of game AI research is heavily skewed towards the development of specific \textit{skills}, leaving the development and analysis of specific \textit{behaviors} an open challenge tackled by few~\cite{de2020predictive,devlin2021navigation}.}

\ian{We propose a non-parametric two-sample hypothesis testing framework to evaluate the behavioral similarity between any two given agents.
\iann{Motivated by our central hypothesis, which we later detail in Section~\ref{sec:method}, we demonstrate the efficacy of this framework within the context of evaluating human-likeness.}}
The importance of this problem stems from its connection to intelligence---a ubiquitous concept often with interpretations conflated across various fields of study.
Expressions of intelligent behavior are so diverse and pervasive that even categorizing its associated abilities into a formal taxonomy is non-trivial.
\ian{As such, researchers interested in evaluating or emulating human-like behavior often focus on a single ability.
While we believe our framework has potential impact across a variety of abilities associated with intelligence, we focus our attention primarily on navigation in 3D space, as it is well-studied across a variety of disciplines~\cite{devlin2021navigation}.}

At the core of the contributions of this work is our statistical framework for evaluating the behavioral similarity of any two agents.
Previous works that build automated proxies to evaluate human-likeness use machine learning-based models to classify or analyze navigation behavior~\cite{de2020predictive,devlin2021navigation}. 
However, our framework represents the navigation behavior of an agent as a distribution of its movements to be compared using our non-parametric two-sample statistical hypothesis test detailed in Section~\ref{sec:method}.
\ian{By doing so, our framework is able to discern nuanced differences in navigation behavior by detecting variations in movement patterns.
Thus, when controlling the sensitivity of the test, we are able to rank agents by their behavioral similarity to human players.
Furthermore, in Section~\ref{sec:human_judgement}, we show that using the resulting $p$-value as a measure of similarity aligns with anonymous human judgment of human-like behavior.
Finally, as detailed in Section~\ref{sec:experiments}, we design a 3D virtual world in which we evaluate the navigation behavior of various agents in a procedurally generated environment.}

\section{Related Work}
\label{sec:related_work}

While the focus of this work is on the evaluation of navigation behavior, we first acknowledge the rich history of imitation learning (IL) literature, which is largely motivated by the difficulty in designing reward functions and seeks to learn a policy from an expert demonstrator using techniques such as behavioral cloning~\cite{zheng2021imitation}.
While significant progress has been made in stabilizing the performance of IL techniques, the driving factors for emulating human-like navigation behavior still remains an open question widely studied across fields ranging from gaming~\cite{alonso2020deep, soni2008bots} to biology~\cite{de2020predictive,banino2018vector}.

Standard approaches to evaluating human-like navigation behavior often require either expert human judges that rely heavily on time-intensive manual efforts or domain-specific metrics that fail to capture fine-grained details of human-likeness \ian{(\textit{e.g.}, monitoring unusual circular movements, detecting collisions, or measuring the cost of detours)~\cite{karpov2013believable, kirby2009companion}.}
Consequently, there has been growing interest in designing automated proxies to use machine learning to accurately evaluate human-likeness~\cite{de2020predictive, devlin2021navigation}.
In their study of navigation behavior, de Cothi \textit{et al.}~\cite{de2020predictive} train a support vector machine to classify navigation patterns of artificial agents and use the resulting model to evaluate similarities with humans and rats.
Following this work, Devlin \textit{et al.}~\cite{devlin2021navigation} take an important step in the scalable analysis of human-like navigation by designing an automated Navigational Turing Test (ANNT) to emulate human judgment using deep neural networks (DNNs).
The outcome of their work is a detection model that takes as an input a representation of navigation behavior and produces as an output a probability that the sample is human.
Of the many interesting results reported in these prior works, they both highlight that proficiency alone is not sufficient for mimicking or measuring human-like behavior.
Motivated by these works, \ian{we provide a new perspective on the evaluation of human-like navigation behavior.
By representing navigation behavior as a distribution of movements, our framework evaluates the behavioral similarity of any two given agents by estimating the differences in their movement patterns.}
To do so, our hypothesis test is motivated by the PT-MMD~\cite{potapov2019pt} framework, which is designed to combine the use of permutation tests (PT) with maximum mean discrepancy (MMD) for the evaluation of generative adversarial networks.
\ian{While this framework has been shown to be effective in evaluating the performance of generative computer vision-based models, we extend it from the computer vision domain into the domain of 3D navigation.}

\section{Preliminaries}
\label{sec:prelims}

Here, we provide an overview of the core techniques used in this work.

\subsection{Deep Reinforcement Learning}
\label{sec:rl}

We assume the standard formulation of reinforcement learning as a Markov decision process in which an agent interacts with an environment $\mathcal{E}$ according to a policy with the goal of maximizing its cumulative expected reward.
Let $\pi_\theta(a | s)$ denote the decision policy of a given agent, where $\pi_\theta$ models the conditional distribution over action $a$ given state $s$ and is parameterized by $\theta$.
In deep reinforcement learning, this conditional distribution is modeled using a deep neural network (DNN).
At each time step $t$, the agent observes the current state $s_t$ and samples an action $a_t$ according to $\pi_\theta$.
The environment $\mathcal{E}$ then responds with a scalar reward $r_t$ that reflects the value of that transition and a new state $s_{t+1}$, which is sampled from the transition dynamics of $\mathcal{E}$ denoted by $p_{\mathcal{E}}(s_{t+1}|s, a)$.
Note that, given state $s$ and action $a$, the transition dynamics satisfy the Markov property such that the probability distribution over the next state is conditionally independent of all previous states and actions.
A sequence of state-action pairs spanning an initial state $s_0$ to a terminal state $s_N$ is often defined as an episode, and a continuous subsequence of an episode over a horizon of $T \leq N$ time steps is often defined as a trajectory, which we denote as $\tau$.

\subsection{Maximum Mean Discrepancy}
\label{sec:mmd}

Maximum mean discrepancy (MMD) is a class of kernel-based divergence metrics often used to compute the distance between the projections of two high-dimensional data distributions~\cite{gretton2012kernel}.
In practice, this distance is empirically estimated by independently drawing $n$ samples from each distribution.
Given that kernel $k$ maps to a reproducing kernel Hilbert space, then the MMD distance given by Eq.~\ref{eq:mmd} is zero if and only if the distributions $X$ and $Y$ are identical.
Here, $\mathbf{x}$ and $\mathbf{y}$ denote sample distributions of $n$ elements independently drawn from $X$ and $Y$, respectively, where $\mathbf{x}'$ indicates that the sample has been shuffled and $\EX_{X,Y}[k(\mathbf{x}, \mathbf{y})]$ is estimated either pairwise or element-wise.
Estimating $\EX_{X,Y}[k(\mathbf{x}, \mathbf{y})]$ with pairwise distances such that  $\EX_{X,Y}[k(\mathbf{x}, \mathbf{y})] = \frac{1}{n^2} \sum_{i=1}^n \sum^n_{j=1} k(x_i, y_j)$ uses each sample point to maximum effect and has been shown to yield better results~\cite{gretton2012kernel}.
Therefore, in our experiments, we use the pairwise estimation of $\EX_{X,Y}[k(\mathbf{x}, \mathbf{y})]$ with the standard Gaussian kernel and the Euclidean distance function such that $k(\mathbf{x}, \mathbf{y}) = -\exp({\Vert \mathbf{x} - \mathbf{y} \Vert^2}) / 2 \sigma^2 $.
{Here,} $\sigma$ is often referred to as the \textit{kernel bandwidth} and is commonly set to the median pairwise distance of the aggregated samples from $X$ and $Y$~\cite{gretton2012kernel}.

\vspace{-0.3cm}
{
\footnotesize
\begin{equation}
	\text{MMD}_k[\mathbf{x}, \mathbf{y}] = \EX_{X, X}[k(\mathbf{x}, \mathbf{x}')] + \EX_{Y, Y}[k(\mathbf{y}, \mathbf{y}')] - 2 \EX_{X, Y}[k(\mathbf{x}, \mathbf{y})]
	\label{eq:mmd}
\end{equation}
}

\vspace{-0.6cm}
\subsection{Non-parametric Statistical Hypothesis Testing}
\label{sec:bootstrap}

Statistical inference is the process of drawing conclusions about a population parameter or population distribution from sample distributions.
Unlike parametric statistical inference, which estimates sample statistics to model a population distribution using assumptions about the data, non-parametric statistical inference analyzes the sample distribution directly.
Among this class of tests and tools, sampling methods such as bootstrap resampling and permutation testing have gained particular attention owing to their increased performance and generality over traditional methods without requiring distributional assumptions~\cite{hesterberg2005boostrap}.
With bootstrap resampling, samples of size $m$ are repeatedly drawn {with replacement} from a sample distribution of size $n$ to recompute a sample statistic~\cite{efron1982jackknife}.
Although~\cite{efron1982jackknife} originally proposed that $m=n$, it has been shown that the $m$ out of $n$ bootstrapping (\textit{i.e.}, $m \leq n$) can yield more consistent results~\cite{bickel2001bootstrap}.
Similar to bootstrap resampling, permutation tests do not require \textit{a priori} knowledge of the data distribution; however, unlike bootstrap resampling, this class of tests resamples from the sample distribution \textit{without replacement} and is often used for hypothesis testing.
\ian{Recent work has combined the use of permutation test (PT) with MMD to design a statistical framework for the evaluation of generative models~\cite{potapov2019pt}.
However, the statistical strength of their test precludes its use as a measure of similarity as the resulting $p$-values are almost always 0.
Thus, we construct our test using $m$ out of $n$ bootstrap resampling for the purpose of controlling sensitivity when rank ordering agents by behavioral similarity, as further discussed in Section~\ref{sec:method}.}

\section{The Human-like Behavior Hypothesis Test}
\label{sec:method}

To analyze the similarity in the behavior of human players to that of artificial agents, we design a non-parametric two-sample hypothesis test that can be used to compare the navigation \ian{behavior of} any two agents.
To construct such a test, we first \ian{use episodic trajectories collected from an agent to represent their navigation behavior} as a distribution of movements, as detailed in Section~\ref{sec:topdown}.
We then compare the distributions of movements \ian{derived from two agents} using our test detailed in Section~\ref{sec:test}.
Finally, the $p$-value \ian{of our proposed test} can be used as a measure of similarity between the two distributions, as shown in Section~\ref{sec:metric}.
To motivate such a test, we first articulate our central hypothesis: \\

\noindent \textbf{The Behavioral Similarity Hypothesis.} \textit{The behaviors of any two agents are sufficiently similar if the distributions over their respective trajectories are sufficiently similar.} \\

\ian{Note that our central hypothesis is agnostic to the nature of the agent; however, to contextualize the details our test, we consider the setting in which our framework is used to evaluate the human-likeness of a reinforcement learning-based (RL-based) artificial agent.}
In such a setting, let $\pi_\theta$ denote the policy used to drive this agent to complete a given task when deployed in environment $\mathcal{E}$.
Similarly, let $\pi^*$ denote the human player that is independently given control over the same agent to complete the same task in the same environment $\mathcal{E}$.
As such, we assume that both the RL-based agent and the human player are bound to the same initial state distribution $p(s_0)$ with the same transition dynamics $p_\mathcal{E}(s_{t+1}|s_t, a_t)$.
We denote the distribution of the trajectories induced by decision policy $\pi_\theta$ as $P_\theta$ and the distribution of trajectories induced by the human player as $P^*$.

\subsection{Representing the Navigation Behavior of Agents}
\label{sec:topdown}

As discussed in Section~\ref{sec:rl}, an episode is defined by a sequence of state-action pairs spanning initial state $s_0$ to terminal state $s_N$.
\ian{To represent the navigation behavior from each episode, we use the absolute 3D location of the agent at each time step, as it is a minimal representation of an agent's response to a given state.}
We then transform each episode into a distribution of movements by subsampling fixed-length trajectories uniformly {with} replacement from each episode using a finite time horizon denoted by $T$.
More formally, let $c_t$ be the 3-dimensional Cartesian coordinates of an agent at time $t$ and let $\mathbf{c}(\tau)$ be the sequence of coordinates for a given trajectory $\tau$ such that $\mathbf{c}(\tau) = \{c_0, \cdots, c_T \}$ and $c_t = (x_t, y_t, z_t)$.
Given an episode of length $N$, we consider overlapping trajectories to be uniquely different such that $\mathbf{c}(\tau_i)$ and $\mathbf{c}(\tau_j)$ have the same probability $\frac{1}{N -T}$ of being sampled, where  $\tau_i = \{s_0, \cdots, s_T\}$, $\tau_j = \{s_1, \cdots, s_{T+1}\}$, and $T < N$.
To ensure that we are analyzing movement without being biased by absolute location, we subtract the initial Cartesian coordinate $c_0$ from each sample $\mathbf{c}(\tau)$ so that each movement starts from the origin.

\ian{When sample distributions are generated over a set of episodes collected from a given agent, we independently subsample $K$ trajectories of length $T$ from each episode {with} replacement.}
\ian{When evaluating agents deployed in large, dynamic environments, we observe that the number of time steps taken to complete even simple tasks is heavily skewed right.
Thus, to correct for any biases from larger episodes with more time steps, we set $K$ to the length of the largest episode in the given set.
As such, the size of the sample distribution will be $M \cdot K$, where $M$ is the number of episodes in the set.
This ensures that a trajectory drawn at random from the aggregated sample distribution has a uniform probability of being from any of the episodes collected.} \\

\subsection{Evaluating Behavioral Similarity using Hypothesis Testing}
\label{sec:test}

\ian{Our central hypothesis posits that the similarity between the behaviors of any two given agents can be estimated by the similarity between their respective distributions of trajectories.}
To test this, we consider the setting in which sample distributions $X$ and $Y$ are independently drawn from distributions $P^*$ and $P_\theta$, respectively.
For the purpose of evaluating their behavioral similarity, we evaluate the null hypothesis ($H_0$) that these distributions are equal, against the alternative hypothesis ($H_1$) that they are different, as summarized below.
Note that there is no point-to-point correspondence between $X$ and $Y$.
\begin{align*}
	H_0 &: P^* = P_\theta \\
	H_1 &: P^* \neq P_\theta
\label{eq:hypothesis_test}
\end{align*}

Our test is motivated by the following insight: if the null hypothesis is true, then any difference between $P^*$ and $P_\theta$ should be due to sampling error.
Therefore, we use MMD (Section~\ref{sec:mmd}) and $m$ out of $n$ bootstrap resampling (Section~\ref{sec:bootstrap}) to calculate our test statistic for the purpose of evaluating the difference between sample distributions $X$ and $Y$.
To derive our $p$-value, we evaluate and compare this distribution of MMD distances in two settings: separated and pooled sample distributions.

First, we consider the setting in which we evaluate over separated distributions $X$ and $Y$.
Given that $\mathbf{x}_i$ and $\mathbf{y}_i$ each denote subsamples of size $m$ that are independently drawn with replacement from $X$ and $Y$, respectively, we form a distribution of MMD distances by repeatedly recomputing MMD$_k[\mathbf{x}_i, \mathbf{y}_i]$ over $S$ iterations where $i \in \{1, \cdots, S\}$.
We refer to this distribution of distances as $\delta_{X,Y}$.
Our test statistic, which we denote as $\delta$, is then calculated using Eq.~\ref{eq:test_statistic_final}, where $\text{quantile}(\delta_{X,Y},\alpha)$ returns the $\alpha$-th quantile over the distribution $\delta_{X,Y}$ and $\alpha$ is a hyperparameter designed to control the sensitivity of the test, as discussed further in Section~\ref{sec:metric}.

\begin{equation}
	\delta = \text{quantile}(\delta_{X,Y}, \alpha) \text{ where } \alpha \in (0, 1)
	\label{eq:test_statistic_final}
\end{equation}

Next, we combine sample distributions $X$ and $Y$ to create a pooled sample distribution, which we will refer to as $Z$.
To evaluate in this setting, we once again form a distribution of MMD distances by repeatedly recomputing MMD$_k[\mathbf{x}_i, \mathbf{y}_i]$ over another $S$ independently drawn samples where $i \in \{1, \cdots, S\}$; however, in this setting, $\mathbf{x}_i$ and $\mathbf{y}_i$ are both independently sampled from pooled distribution $Z$ with replacement.
We refer to this distribution of estimates as $\delta_Z$.
Finally, to evaluate our null hypothesis, we define our $p$-value as the percentage of estimates greater than our test statistic $\delta$, as shown in Eq.~\ref{eq:p-value}.

\begin{equation}
	p = \frac{\#(\delta_Z > \delta)}{N}
	\label{eq:p-value}
\end{equation}

\begin{figure}
	\centering
	\includegraphics[width=0.7\linewidth]{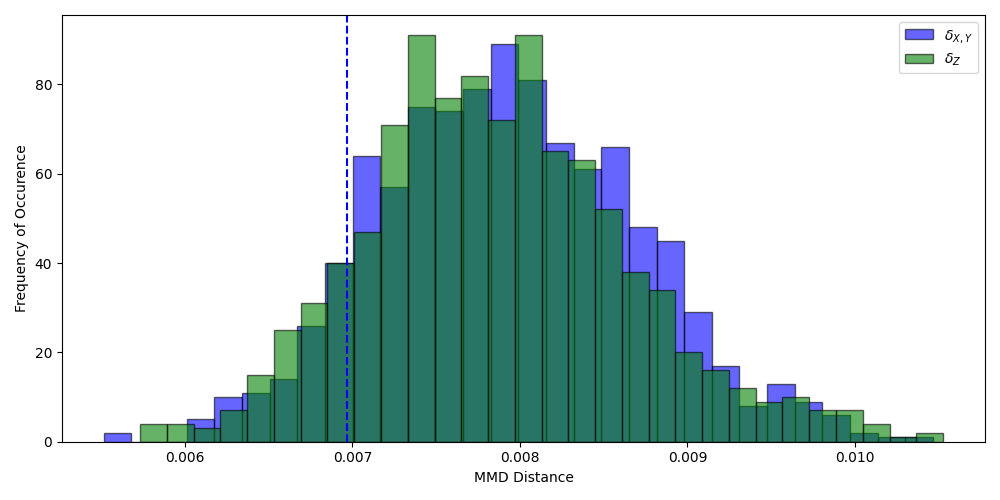}
	\caption{{\small\textbf{Toy Example.} Using a 128-dimensional standard Gaussian distribution, we show that when $P^* = P_\theta$ the resulting $p$-value converges towards $1 - \alpha$ as $S \rightarrow \infty$. In this example, $\alpha = 0.10$, $m=100$, $S = 1000$, and our median observed $p$-value is 88.5\% with an IQR of 0.43\%. Note that the \textcolor{blue}{blue} histogram depicts the distribution of MMD distances as computed over sample distributions $X$ and $Y$ ($\delta_{X,Y}$), while the \textcolor{ForestGreen}{green} histogram depicts the distribution of MMD distances as computed over the pooled sample distribution $Z$ ($\delta_Z$). The test statistic $\delta$ is denoted by the dotted \textcolor{blue}{blue} line.}}
	\label{fig:toy_example_0}
\end{figure}

\begin{figure*}
	\centering
	\subfloat[$P_\theta = \mathcal{N}(0.02, 1)$]{\includegraphics[width=0.2\linewidth]{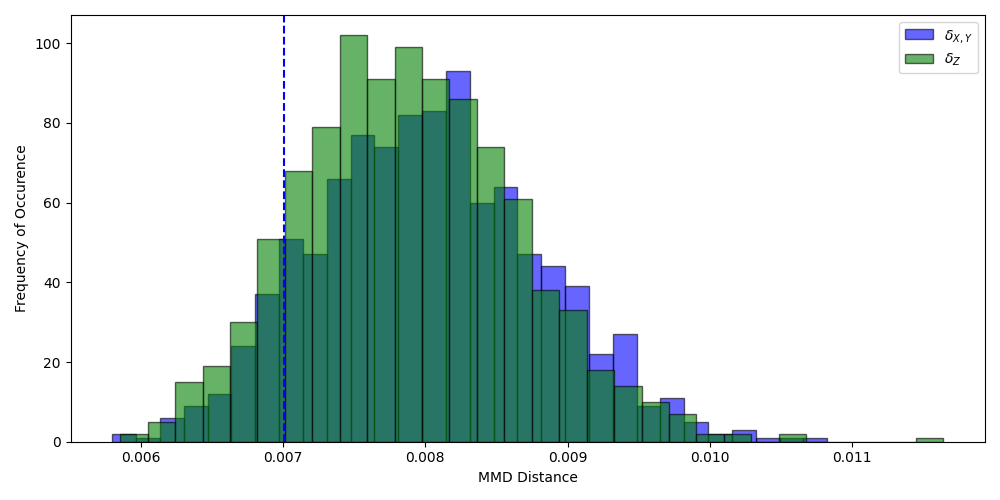}}
	\subfloat[$P_\theta = \mathcal{N}(0.04, 1)$]{\includegraphics[width=0.2\linewidth]{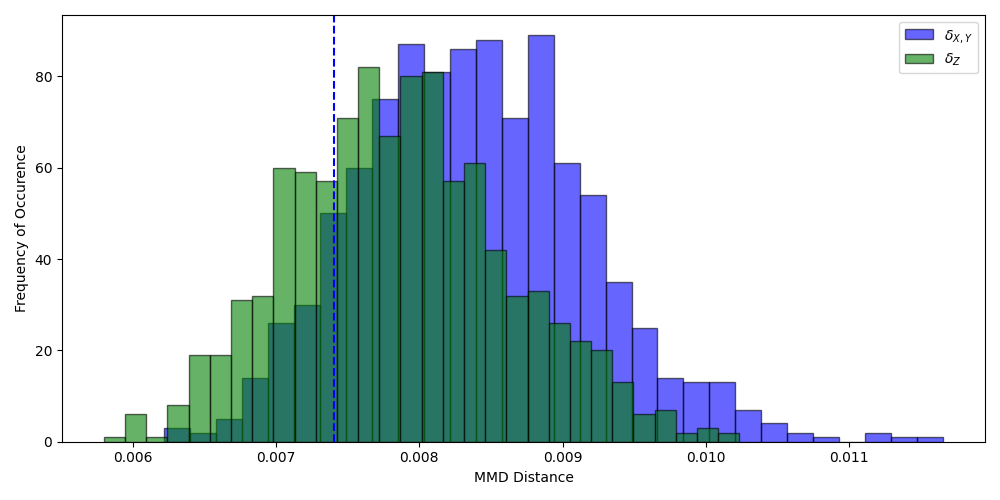}}
	\subfloat[$P_\theta = \mathcal{N}(0.06, 1)$]{\includegraphics[width=0.2\linewidth]{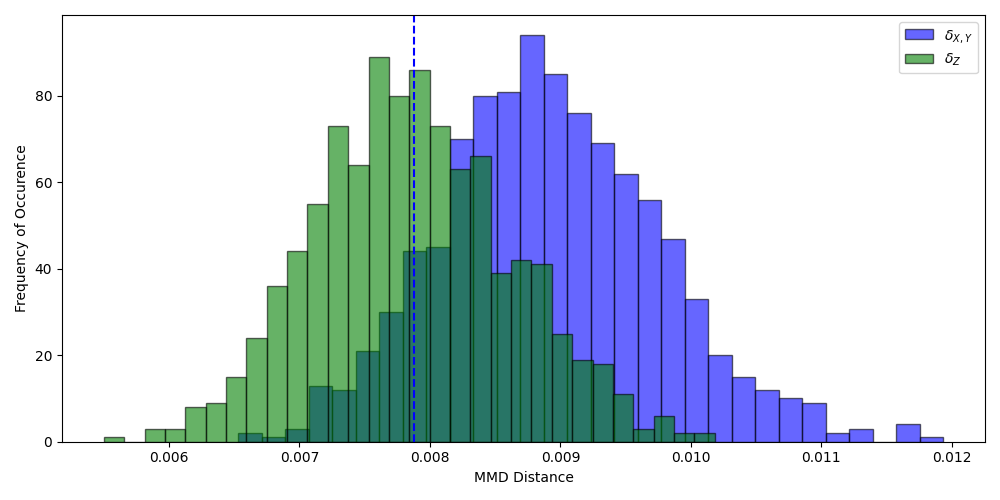}}
	\subfloat[$P_\theta = \mathcal{N}(0.08, 1)$]{\includegraphics[width=0.2\linewidth]{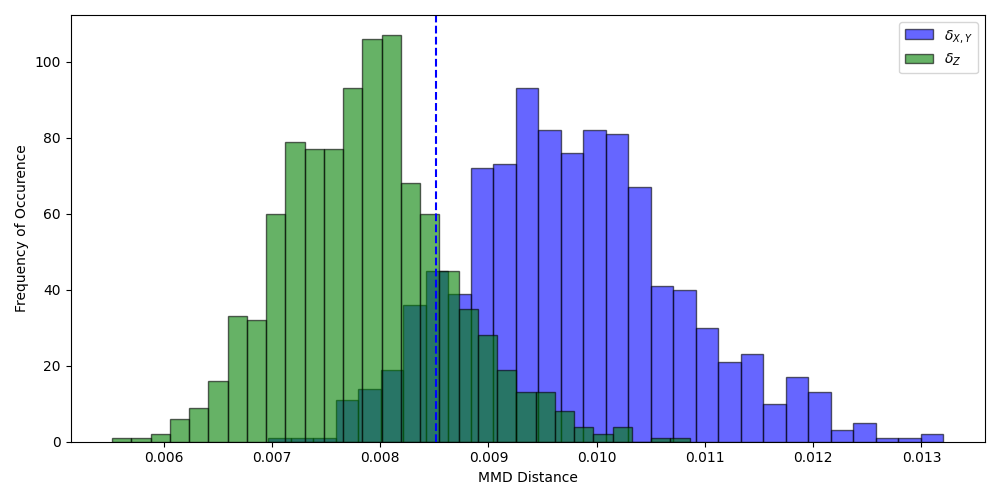}}
	\subfloat[$P_\theta = \mathcal{N}(0.10, 1)$]{\includegraphics[width=0.2\linewidth]{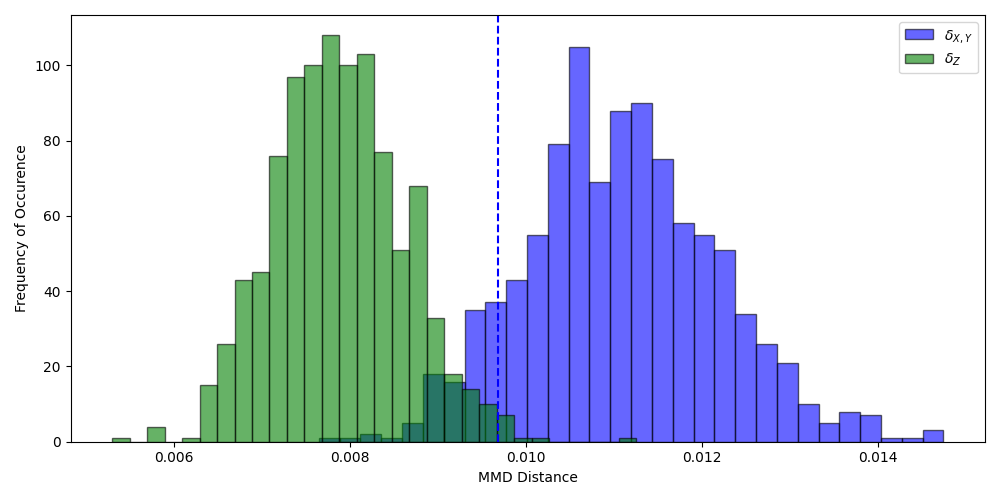}}
	\caption{{\small\textbf{Measuring Similarity in High-Dimensional Distributions.} Given that $P^* = \mathcal{N}(0, 1)$, we incrementally shift $P_\theta = \mathcal{N}(\epsilon, 1)$ by 0.02 to demonstrate the resulting shifts in the distributions of MMD distances as calculated over the separated and pooled sample distributions. For each experiment, we use $m=100$, $\alpha=0.10$, and $S=1000$. The median $p$-values and IQR \ian{over 10 repeats} are reported in Table~\ref{tbl:effect_of_m}.}}
	\label{fig:p-val_examples}
\end{figure*}

\begin{table*}
\centering
\begin{tabular}{|lcccccc|}
\hline
\textbf{} & $P_\theta = \mathcal{N}(0, 1)$ & $P_\theta = \mathcal{N}(0.02, 1)$ & $P_\theta = \mathcal{N}(0.04, 1)$ & $P_\theta = \mathcal{N}(0.06, 1)$ & $P_\theta = \mathcal{N}(0.08, 1)$ & $P_\theta = \mathcal{N}(0.10, 1)$ \\ \hline
\multicolumn{1}{|l|}{$\alpha=0.10$} & \multicolumn{1}{c|}{88.5\% (0.43\%)} & \multicolumn{1}{c|}{85.9\% (0.70\%)} & \multicolumn{1}{c|}{74.4\% (4.20\%)} & \multicolumn{1}{c|}{48.6\% (2.23\%)} & \multicolumn{1}{c|}{{18.4\% (1.10\%)}} & \multicolumn{1}{c|}{1.1\% (0.40\%)} \\ \hline
\multicolumn{1}{|l|}{$\alpha=0.25$} & \multicolumn{1}{c|}{71.3\% (1.02\%)} & \multicolumn{1}{c|}{64.8\% (4.05\%)} & \multicolumn{1}{c|}{50.8\% (3.30\%)} & \multicolumn{1}{c|}{24.8\% (3.05\%)} & \multicolumn{1}{c|}{7.1\% (1.22\%)} & \multicolumn{1}{c|}{0.3\% (0.18\%)} \\ \hline
\multicolumn{1}{|l|}{$\alpha=0.50$} & \multicolumn{1}{c|}{46.7\% (1.30\%)} & \multicolumn{1}{c|}{41.0\% (1.35\%)} & \multicolumn{1}{c|}{24.9\% (1.17\%)} & \multicolumn{1}{c|}{8.5\% (2.05\%)} & \multicolumn{1}{c|}{1.2\% (0.28\%)} & \multicolumn{1}{c|}{0.0\% (0.00\%)} \\ \hline
\end{tabular}
\vspace{0.1cm}
\caption{{\small\textbf{Controlling Sensitivity.} Using our 128-dimensional toy example where $P^* = \mathcal{N}(0,1)$, $P_\theta = \mathcal{N}(\epsilon, 1)$, and $\epsilon \in \{0, 0.02, 0.04, 0.06, 0.08, 0.10 \}$, we show that the selection of $\alpha$ can be used to control the sensitivity of the test. When used to rank order agents by behavioral similarity, this sensitivity control can be used to create more informative comparisons. \ian{We run each experiment 10 times and report the median and IQR of the $p$-value.}}}
\label{tbl:effect_of_m}
\end{table*}

\subsection{Using the Hypothesis Test as a Measure of Similarity}
\label{sec:metric}

Given that $P^*$ and $P_\theta$ are the same distribution under our null hypothesis ($H_0$), then the distribution of MMD distances as computed over the separated sample distributions $X$ and $Y$ should be the same as the distribution of MMD distances as computed over the pooled sample distribution $Z$.
Thus, when $P^* = P_\theta$, it follows that the resulting $p$-value converges towards $1 - \alpha$ as $S \rightarrow \infty$.
{However, when $P^* \neq P_\theta$, we can interpret the derived $p$-value as a measure of closeness between distributions $P^*$ and $P_\theta$.}
To demonstrate this, we consider a series of experiments using a toy example where $P^*$ is distributed as a 128-dimensional standard Gaussian with zero mean and unit variance, which we'll denote as $\mathcal{N}(0, 1)$.
For each experiment, we run $S = 1000$ iterations using subsample size $m = 100$.
Due to the inherent stochasticity of our test, we repeat each experiment 10 times and report both the median $p$-value and the observed interquartile range (IQR).

First, we consider the case where $P_\theta = P^* = \mathcal{N}(0, 1)$ using $\alpha = 0.10$.
We observe a median $p$-value of 88.5\% with an IQR of 0.43\%.
In Fig.~\ref{fig:toy_example_0}, we show the observed distributions of MMD distances as independently calculated over the separated and pooled sample distributions.
Next, to demonstrate how the $p$-value resulting from our hypothesis test can be used as a measure of similarity between distributions $P^*$ and $P_\theta$, we show that $p$ monotonically decreases as the distribution $P_\theta$ is incrementally shifted by epsilon $\epsilon = 0.02$.
We compare each $P_\theta = \mathcal{N}(\epsilon, 1)$ to $P^* = \mathcal{N}(0,1)$ and report the observed results in Table~\ref{tbl:effect_of_m}.
Furthermore, we demonstrate how the distribution of the MMD distances shifts with $P_\theta$ in Fig.~\ref{fig:p-val_examples}.
Note that when $P_\theta = \mathcal{N}(0.10, 1)$, the distributions of $\delta_{X,Y}$ and $\delta_Z$ are nearly separated.
Finally, to control the sensitivity of the test, we can adjust $\alpha$.
Intuitively, higher values of $\alpha$ yield a larger test statistic $\delta$ and therefore increase the sensitivity of our test.
As shown in Table~\ref{tbl:effect_of_m}, a more sensitive test yields lower $p$-values as the distributions diverge.
Therefore, in cases where the behavior of multiple agents are being compared against the behavior of a human player, $\alpha$ can be selected to control for sensitivity and create a more informative comparison.

\section{Experimental Analysis}
\label{sec:experiments}

To evaluate the efficacy of our hypothesis test when measuring similarities in navigation behavior, we design a virtual 3D world in which we study three types of agents deployed to complete a navigation task in a controlled environment.

\subsection{About Our \ian{Environment}}
\label{sec:world}

\begin{figure}
	\centering
	\subfloat[]{\includegraphics[width=0.45\linewidth]{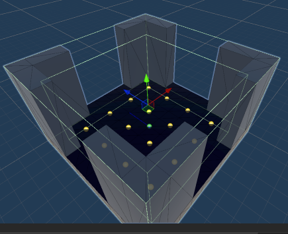}} ~
	\subfloat[]{\includegraphics[width=0.44\linewidth]{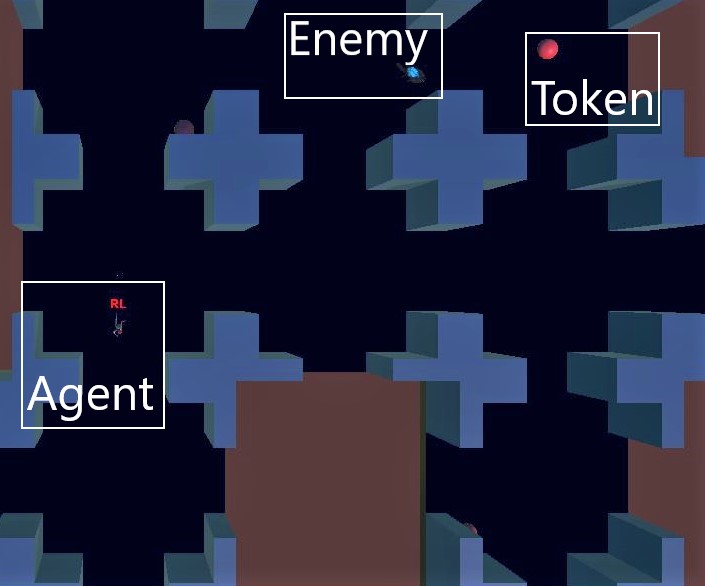}}
	\caption{{\small\textbf{Our 3D Maze Environment.} We design a 3D virtual world in which we study the navigation behavior of various agents. On the left, we show the primitive segment used to procedurally generate random maze-like environments, as shown on the right. Note that the platform is colored in \textcolor{NavyBlue}{dark blue}, while \ian{off-platform} (\textit{i.e.,} falling off the map) is colored in \textcolor{brown}{brown}.}}
	\label{fig:maze_env}
\end{figure}

Our \ian{environment} is governed by the following constants: all agents have the same movement dynamics; all simulated objects are bounded by similar physical properties; and the set of topological building blocks are limited to a set of primitives.
Each rendered environment is divided into equal-sized segments defined by a platform with a set of walls around each corner such that randomly generating any number of segments yields a maze-like environment, as shown in Figure~\ref{fig:maze_env}.
Each segment is sub-divided into $M$ spawn points, each of which can spawn a token with uniform probability.
Each environment is randomly generated to have $N$ segments, where each segment generates a single token with probability $p_\text{token}$ in one of its $M$ positions such that, for any given segment, a spawn point will generate a token with probability $\frac{p_\text{token}}{M}$.
These spawn points similarly govern the procedural generation of enemies, where the spawn point of any given segment will generate an enemy with probability $\frac{p_\text{enemy}}{M}$.
Enemies have a visual radius of roughly 5 segments and are controlled using navigation meshes (NavMesh)~\cite{snook2000navmesh} to target and attack the agent without collaboration.
In our experiments, $p_\text{token} = 75\%$, $p_\text{enemy} = 25\%$, $M=16$, and $N$ is sampled from a discrete uniform distribution between 5 and 30 such that $N \sim U(5, 30)$.
At the start of each episode, an agent is deployed to collect tokens that are randomly scattered around the rendered maze.
To successfully complete the navigation task, an agent must collect all of the tokens available within the rendered environment without falling off the map or being attacked by an enemy.
Note that all generated segments are connected.
An episode ends either when all tokens are collected or if the maximum step count is reached\footnote{\ian{We use this maximum step count primarily for training our RL-based agent. After fully training the RL-based agent, we observe that the maximum step count is never reached by any of our agents.}}.
If an agent falls off the map or is attacked by an enemy, the agent is respawned to continue.

\subsection{About Our Agents}
\label{sec:agents}

\begin{figure*}
	\centering
	\includegraphics[width=0.8\linewidth]{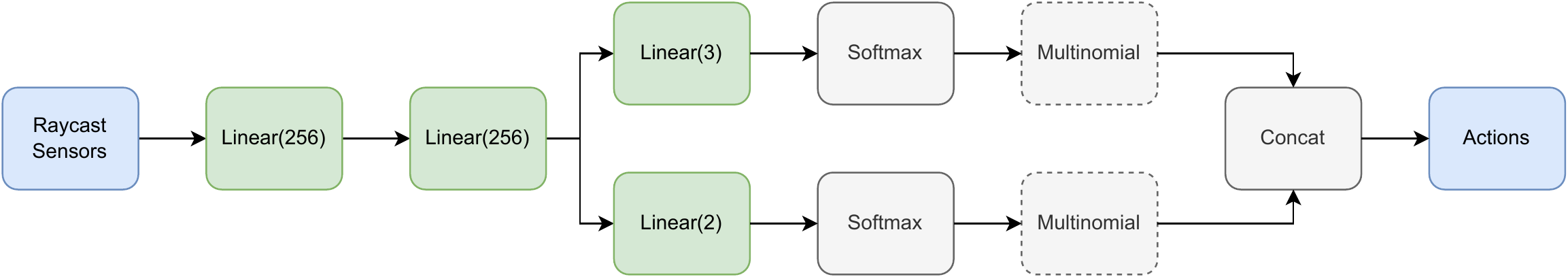}
	\caption{{\small\textbf{Deep Neural Network (DNN) Architecture for RL-based Agent.} Our RL-based agent uses raycast sensors to form its observation space and produces discrete actions as an output. Here, the \textcolor{blue}{blue} blocks denote our input and output, \textcolor{ForestGreen}{green} blocks denote learnable blocks, and \textcolor{gray}{grey} blocks denote static {non-learnable functions}. The dotted line around the multinomial block denotes that it is frozen during inference, at which point we greedily take the action that maximizes the expected return as {determined} by the preceding softmax block.}}
	\label{fig:dnn_arch}
\end{figure*}

We study the behavior of the following three types of agents: human players, RL-based, and NavMesh-based.
All agents are rendered as a humanoid character with physics-based movement dynamics that define their action space $\mathcal{A}$.
At each time step, an agent has two sets of \ian{independent} actions: (1) the agent can move forward or not; and (2) the agent can move left or right or not.
Note that this yields actions such as ``Forward-Left" or ``None-Right". \\

\noindent\textbf{Human players} are controlled using {WASD keys} and are bounded to the same action space $\mathcal{A}$ as both RL-based and NavMesh-based agents; however, unlike RL-based or NavMesh-based agents, the observation space of the human player is a first-person point of view of the humanoid agent. In our experiments, we evaluate human players using the same reward function as defined for RL-based agents. \\

\noindent\textbf{RL-based agents} are controlled by policy $\pi_\theta$ modeled by the DNN depicted in Fig.~\ref{fig:dnn_arch}.
The agent is trained using a curriculum in which we monotonically increase the maximum number of segments ($N_\text{max}$) from which to sample $N$ from.
Given that $N \sim U(5, N_\text{max})$, we increase $N_\text{max}$ by 2 every 1M epochs starting from $N_\text{max} = 10$ and finishing with $N_\text{max} = 30$ after 10M epochs.
The learnable parameters are optimized using proximal policy optimization (PPO)~\cite{schulman2017proximal}.
To form its observation space, the agent is equipped with several raycast sensors, each cast from the agent's head to determine the observation vector that defines the current state $s_t$.
In our experiments, we use a simple reward signal where the agent receives +1 for collecting a token, -1 for either falling off the map or being attacked by an enemy, and 0 otherwise. \\

\noindent\textbf{NavMesh-based agents} are controlled using a behavior tree that knows where all rendered tokens are in the environment and greedily collects each according to proximity.
Unlike the RL-based {agents}, these agents do not use raycast sensors but rather use a NavMesh~\cite{snook2000navmesh} to guide them through the environment. {They have the same action space and are evaluated with the same reward function as RL-based {agents}.} \\

\begin{table*}
\centering
\begin{tabular}{|lcccccc|}
\hline
\textbf{} & \textbf{\% Coins} & \textbf{\% Reward} & \textbf{\# Falls} & \textbf{\# Attacks} & \textbf{Steps / Seg.} & \textbf{Walls / Seg.} \\ \hline
\multicolumn{1}{|l|}{Agent 1} & \multicolumn{1}{|c|}{\textbf{100.0\%}}      & \multicolumn{1}{c|}{\textbf{96.6\%}} & \multicolumn{1}{c|}{$0.03$} & \multicolumn{1}{c|}{\textbf{0.40}} & \multicolumn{1}{c|}{$189.89$} & \multicolumn{1}{c|}{0.75} \\ \hline
\multicolumn{1}{|l|}{Agent 2} & \multicolumn{1}{|c|}{96.0\%}      & \multicolumn{1}{c|}{{89.5\%}} & \multicolumn{1}{c|}{0.18} & \multicolumn{1}{c|}{{0.60}} & \multicolumn{1}{c|}{287.86} & \multicolumn{1}{c|}{0.81} \\ \hline
\multicolumn{1}{|l|}{Agent 3} & \multicolumn{1}{|c|}{\textbf{100.0\%}} & \multicolumn{1}{c|}{$85.4\%$} & \multicolumn{1}{c|}{\textbf{0.00}} & \multicolumn{1}{c|}{$1.55$} & \multicolumn{1}{c|}{\textbf{163.09}} & \multicolumn{1}{c|}{\textbf{0.16}} \\ \hline
\multicolumn{1}{|l|}{Agent 4} & \multicolumn{1}{|c|}{\textbf{100.0\%}} & \multicolumn{1}{c|}{{92.6\%}} & \multicolumn{1}{c|}{\textbf{0.00}} & \multicolumn{1}{c|}{{0.58}} & \multicolumn{1}{c|}{221.53} & \multicolumn{1}{c|}{0.46}  \\ \hline
\end{tabular}
\vspace{0.1cm}
\caption{{\small\textbf{Which agents are the most human-like?}
\ian{We demonstrate that domain-specific metrics are often too coarse-grained to evaluate and compare the human-like navigation behavior of agents.
Here, agents 1 and 4 are human players, agent 2 is our RL-based agent trained over 10M epochs using PPO~\cite{schulman2017proximal}, and agent 3 is the NavMesh-based agent.
Note that these metrics, which are discussed in Section~\ref{sec:evaluation}, are aggregated over the same 40 procedurally generated mazes for each agent.}}}
\label{tbl:maze_agents}
\end{table*}

\begin{figure}
	\centering
	\subfloat[]{\includegraphics[width=0.3\linewidth]{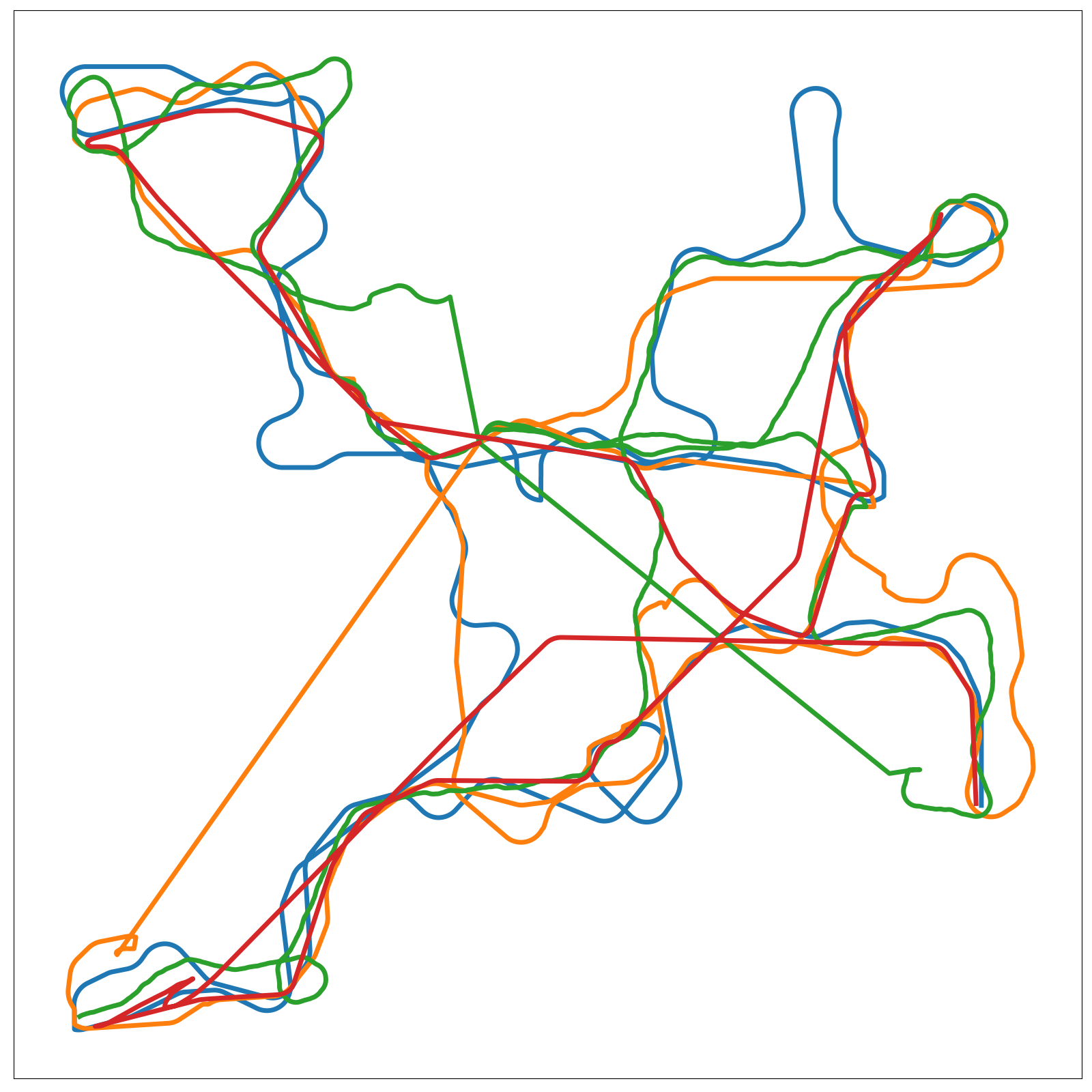}} ~~
	\subfloat[]{\includegraphics[width=0.3\linewidth]{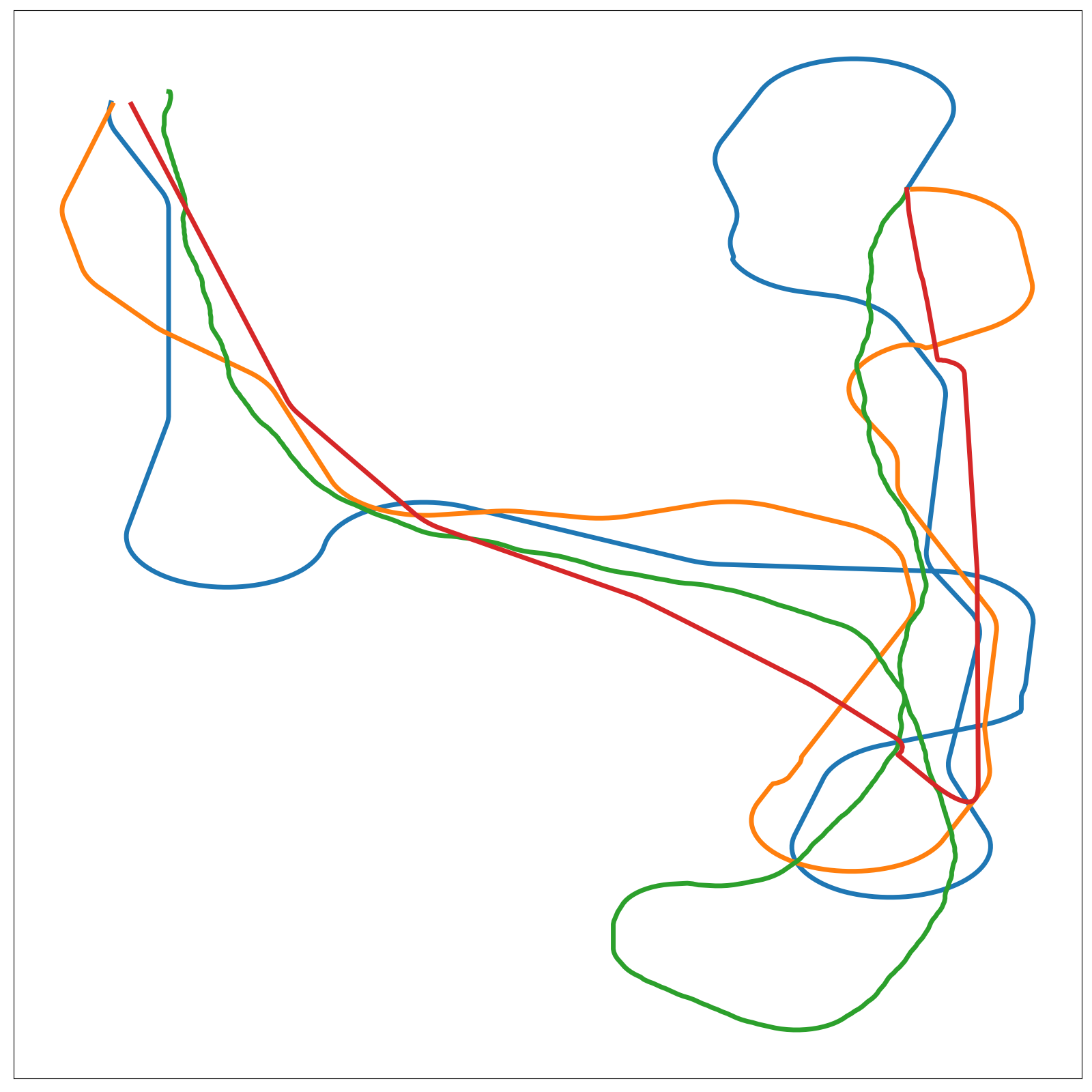}} \\
	\subfloat[]{\includegraphics[width=0.3\linewidth]{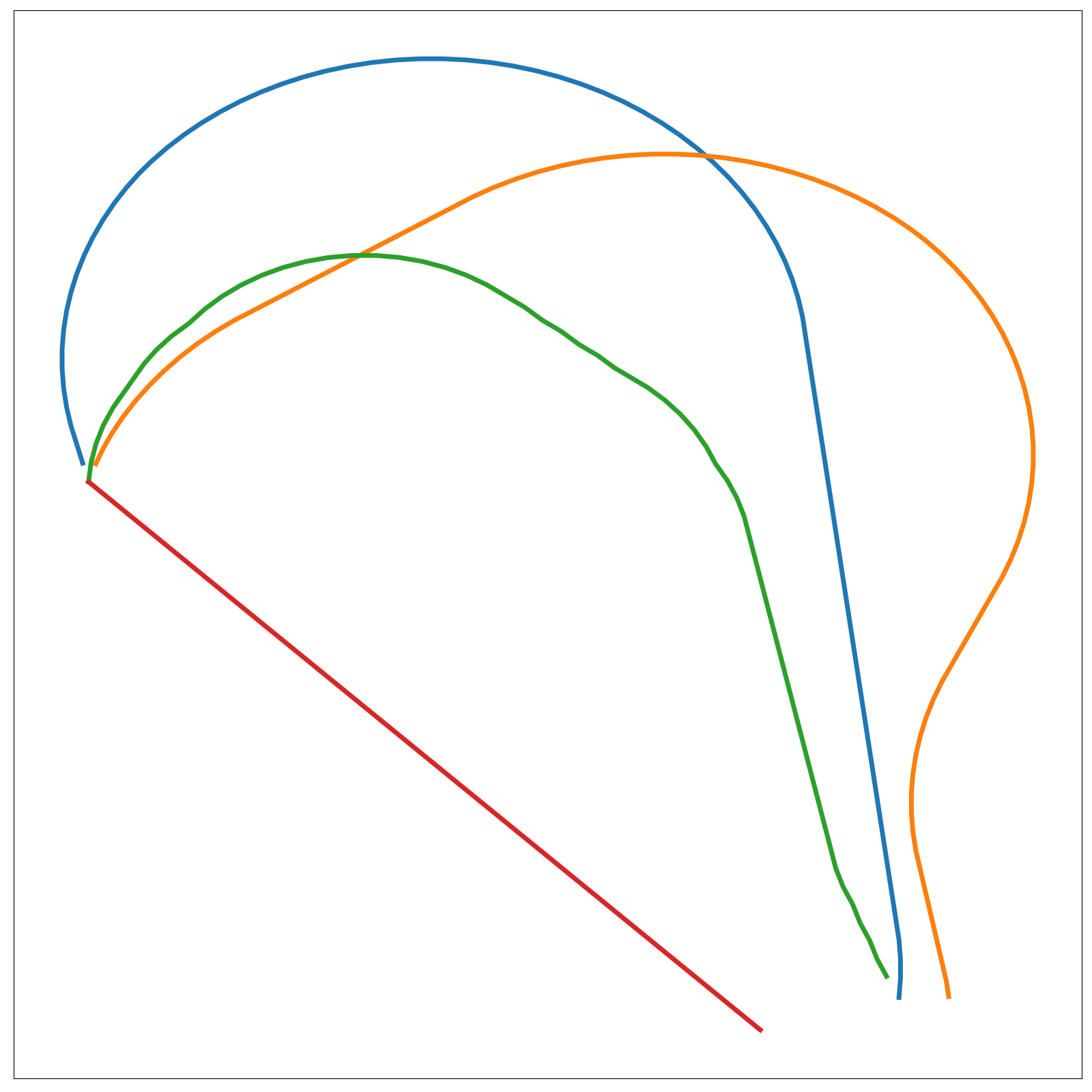}} ~~
	\subfloat[]{\includegraphics[width=0.3\linewidth]{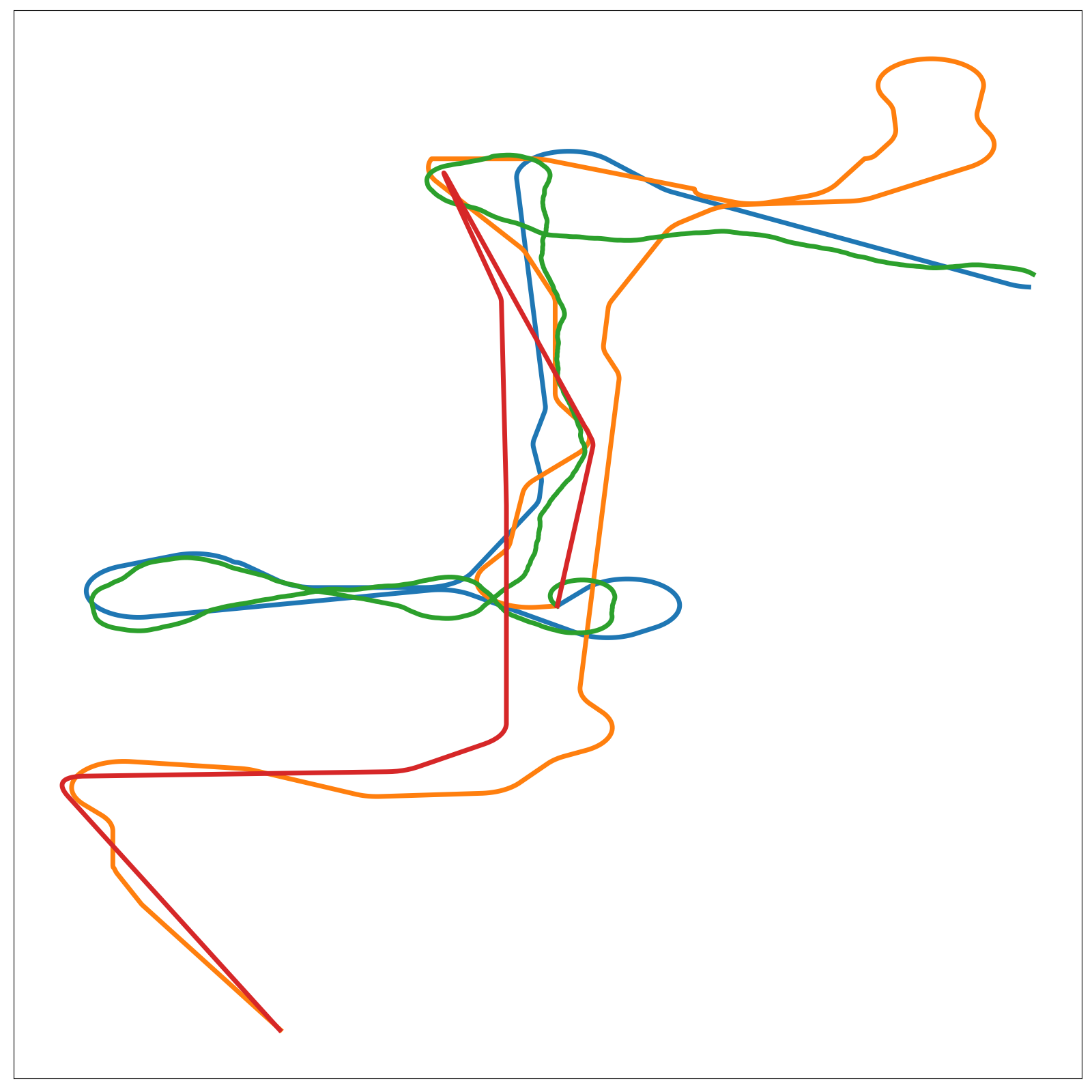}}
\caption{{\small\textbf{Examples of Trajectories from Various Agents.} Here, we demonstrate the varied complexity of the navigation task by visualizing the navigation behavior of each agent over four different episodes. Note that each episode uses a different random seed to randomly generate a different environment. Here, \textcolor{blue}{blue} and \textcolor{BurntOrange}{orange} denote two unique human players, while \textcolor{ForestGreen}{green} and \textcolor{red}{red} denote the RL-based and NavMesh-based agents, respectively.}}
\label{fig:trajectories_agents}
\end{figure}

\subsection{Evaluating the Human-Like Behavior of Artificial Agents}
\label{sec:evaluation}

\ian{Previous work has shown that RL-based agents exhibit more human-like navigation behavior than NavMesh-based counterparts~\cite{alonso2020deep}.
Qualitatively, we observe this to hold true with our agents.
To quantify these differences, we evaluate each agent using both domain-specific metrics as well as our human-like behavior hypothesis test.
First, to visualize the complexities of the navigation behaviors of each agent, we show four examples of trajectories in Fig.~\ref{fig:trajectories_agents} visualized using the ``topdown" projection proposed in~\cite{devlin2021navigation}.
Note that even when completing our simple task of collecting tokens within a randomly generated maze, the navigation behavior of each agent varies greatly even within the same environment.}

It is often the case that domain-specific approaches to analyze behavioral similarity use metrics designed to evaluate an agent's proficiency in a given task; however, proficiency alone is not sufficient to discern human-like behavior.
To demonstrate the pitfalls of domain-specific approaches, we use the following standard domain-specific metrics to evaluate the performance of each of our agents: the percentage of coins collected (\% Coins), percentage of the maximum reward received (\% Reward), the average number of falls off the map (\# Falls), the average number of attacks by an enemy (\# Attacks), the average number of steps taken per segment (Steps / Seg.), and the average number of walls hit per segment (Walls / Seg.).
\ian{To ensure an even comparison, each agent is evaluated over the same 40 mazes.}
We report the observed results in Table~\ref{tbl:maze_agents} and purposefully obfuscate the nature of each agent to ask the reader to initially classify them before reading the caption.
We find that while domain-specific metrics provide a useful insight into the navigation skills of a given agent, they are unable to capture the nuanced aspects of human-likeness that are necessary for {evaluating behavior}.
Therefore, to measure the behavioral similarity of these agents, we use our hypothesis test detailed in Section~\ref{sec:method}.

For each experiment with this environment, we run $S=1000$ iterations with a subsample size $m = 1000$.
Similar to our experiments in Section~\ref{sec:human_judgement}, we first evaluate the human-to-human likeness as a baseline; however, unlike the previous experiments, we are able to separate human data into two individual human players.
Therefore, we evaluate the two humans against each other, where each sample dataset is aggregated over all 40 episodes using the techniques detailed in Section~\ref{sec:topdown}.
As shown in Table~\ref{tbl:our_environment}, we observe stable results across all selections of $T$ and $\alpha$.
Next, we apply our hypothesis test to compare the human-like navigation behaviors of both the RL-based and NavMesh-based agents.
For these experiments, we use the aggregated 80 episodes collected from both humans to form our sample set $X \sim P^*$ and evaluate each artificial agent using their trajectories collected over the 40 procedurally generated environments.
In this study, we observe that the longest episodes for the two human players were 10.9k and 7.6k time steps while the longest episodes for the RL-based and NavMesh-based agents were 12.9k and 5.7k, respectively.
As before, we repeat each experiment 10 times and report the median $p$-value and IQR across different time horizons ($T$) and alpha levels ($\alpha$) in Table~\ref{tbl:our_environment}.
We observe that the results from our test consistently align with our intuition that RL-based agents are able to express more human-like navigation behavior than NavMesh-based agents.

\begin{table}
\centering
\begin{tabular}{|c|c|c||c|c|}
\hline
\multirow{4}{*}{\textbf{T = 4}} & {$\alpha$} & \textbf{Human} & \textbf{RL} & \textbf{NavMesh} \\ \cline{2-5} 
                                & \multicolumn{1}{c|}{0.10}   & \multicolumn{1}{c||}{70.1\% (2.35\%)}   & \multicolumn{1}{c|}{74.2\% (2.68\%)}          &                                  \multicolumn{1}{c|}{0.0\% (0.00\%)}   \\ \cline{2-5} 
                                & \multicolumn{1}{c|}{0.25}   & \multicolumn{1}{c||}{48.4\% (2.35\%)}   & \multicolumn{1}{c|}{53.9\% (1.90\%)}          &                                   \multicolumn{1}{c|}{0.0\% (0.00\%)}  \\ \cline{2-5} 
                                & \multicolumn{1}{c|}{0.50}    & \multicolumn{1}{c||}{24.9\% (2.43\%)}    & \multicolumn{1}{c|}{36.3\% (2.15\%)}           &  \multicolumn{1}{c|}{0.0\% (0.00\%)}                                 \\ \hline \hline
\multirow{3}{*}{\textbf{T = 8}} & \multicolumn{1}{c|}{0.10}   & \multicolumn{1}{c||}{69.1\% (2.55\%)}   & \multicolumn{1}{c|}{74.6\% (2.35\%)}          &                                   \multicolumn{1}{c|}{0.0\% (0.00\%)}  \\ \cline{2-5} 
                                & \multicolumn{1}{c|}{0.25}   & \multicolumn{1}{c||}{47.7\% (2.40\%)}   & \multicolumn{1}{c|}{58.5\% (1.55\%)}          &                                   \multicolumn{1}{c|}{0.0\% (0.00\%)}  \\ \cline{2-5} 
                                & \multicolumn{1}{c|}{0.50}    & \multicolumn{1}{c||}{24.7\% (2.78\%)}    & \multicolumn{1}{c|}{36.4\% (2.13\%)}           &                                   \multicolumn{1}{c|}{0.0\% (0.00\%)}  \\ \hline \hline
\multirow{3}{*}{\textbf{T = 16}} & \multicolumn{1}{c|}{0.10}   & \multicolumn{1}{c||}{79.4\% (1.77\%)}   & \multicolumn{1}{c|}{74.7\% (3.05\%)}          &                                   \multicolumn{1}{c|}{0.0\% (0.00\%)} \\ \cline{2-5} 
                                & \multicolumn{1}{c|}{0.25}    & \multicolumn{1}{c||}{62.7\% (1.85\%)}    & \multicolumn{1}{c|}{59.7\% (3.45\%)}           &                                   \multicolumn{1}{c|}{0.0\% (0.00\%)} \\ \cline{2-5}  
                                & \multicolumn{1}{c|}{0.50}    & \multicolumn{1}{c||}{40.0\% (2.53\%)}    & \multicolumn{1}{c|}{35.8\% (3.83\%)}           &                                   \multicolumn{1}{c|}{0.0\% (0.00\%)} \\ \hline \hline
\multirow{3}{*}{\textbf{T = 32}} & \multicolumn{1}{c|}{0.10}   & \multicolumn{1}{c||}{77.3\% (3.13\%)}   & \multicolumn{1}{c|}{78.3\% (2.23\%)}          &                                  \multicolumn{1}{c|}{0.0\% (0.00\%)}   \\ \cline{2-5} 
                                & \multicolumn{1}{c|}{0.25}   & \multicolumn{1}{c||}{62.4\% (1.93\%)}   & \multicolumn{1}{c|}{61.1\% (1.63\%)}          &                                  \multicolumn{1}{c|}{0.0\% (0.00\%)}  \\ \cline{2-5} 
                                & \multicolumn{1}{c|}{0.50}    & \multicolumn{1}{c||}{39.6\% (2.45\%)}    & \multicolumn{1}{c|}{37.1\% (2.53\%)}           &                                   \multicolumn{1}{c|}{0.0\% (0.00\%)} \\ \hline
\end{tabular}
\caption{{\small\textbf{Evaluating human-like navigation behavior.}
We evaluate the similarity of two human players (left) before evaluating the human-like navigation behavior of our RL-based (center) and NavMesh-based (right) agents. Interestingly, we observe that the RL-based agent is more similar to aggregated human players than the individual human players are to each other; however, we also observe that the results when comparing the human-likeness of RL-based and NavMesh-based agents align with our intuition across all selections of $T$ and $\alpha$. Each experiment is repeated 10 times and we report the median and IQR $p$-value.}}
\label{tbl:our_environment}
\end{table}

\section{The Human-like Behavior Hypothesis Test and Human Judgment}
\label{sec:human_judgement}

\begin{table}
\centering
\begin{tabular}{|c|c|c||c|c|}
\hline
\multirow{4}{*}{\textbf{T = 4}} & {$\alpha$} & \textbf{Human} & \textbf{Hybrid} & \textbf{Symbolic} \\ \cline{2-5} 
                                & \multicolumn{1}{c|}{0.10}   & \multicolumn{1}{c||}{90.5\% (0.95\%)}   & \multicolumn{1}{c|}{19.6\% (1.63\%)}          &                                  \multicolumn{1}{c|}{8.6\% (2.28\%)}   \\ \cline{2-5} 
                                & \multicolumn{1}{c|}{0.25}   & \multicolumn{1}{c||}{75.3\% (3.50\%)}   & \multicolumn{1}{c|}{7.0\% (1.48\%)}          &                                   \multicolumn{1}{c|}{2.8\% (0.88\%)}  \\ \cline{2-5} 
                                & \multicolumn{1}{c|}{0.50}    & \multicolumn{1}{c||}{50.9\% (1.93\%)}    & \multicolumn{1}{c|}{1.7\% (0.25\%)}           &  \multicolumn{1}{c|}{0.4\% (0.25\%)}                                 \\ \hline \hline
\multirow{3}{*}{\textbf{T = 8}} & \multicolumn{1}{c|}{0.10}   & \multicolumn{1}{c||}{89.7\% (1.80\%)}   & \multicolumn{1}{c|}{19.7\% (1.43\%)}          &                                   \multicolumn{1}{c|}{8.5\% (0.75\%)}  \\ \cline{2-5} 
                                & \multicolumn{1}{c|}{0.25}   & \multicolumn{1}{c||}{76.7\% (2.75\%)}   & \multicolumn{1}{c|}{6.6\% (1.20\%)}          &                                   \multicolumn{1}{c|}{2.5\% (0.45\%)}  \\ \cline{2-5} 
                                & \multicolumn{1}{c|}{0.50}    & \multicolumn{1}{c||}{50.5\% (1.63\%)}    & \multicolumn{1}{c|}{1.6\% (0.25\%)}           &                                   \multicolumn{1}{c|}{0.5\% (0.18\%)}  \\ \hline \hline
\multirow{3}{*}{\textbf{T = 16}} & \multicolumn{1}{c|}{0.10}   & \multicolumn{1}{c||}{90.4\% (1.38\%)}   & \multicolumn{1}{c|}{20.6\% (3.03\%)}          &                                   \multicolumn{1}{c|}{8.7\% (0.85\%)} \\ \cline{2-5} 
                                & \multicolumn{1}{c|}{0.25}    & \multicolumn{1}{c||}{73.5\% (2.48\%)}    & \multicolumn{1}{c|}{7.6\% (0.97\%)}           &                                   \multicolumn{1}{c|}{2.6\% (0.28\%)} \\ \cline{2-5}  
                                & \multicolumn{1}{c|}{0.50}    & \multicolumn{1}{c||}{49.8\% (2.87\%)}    & \multicolumn{1}{c|}{1.7\% (0.45\%)}           &                                   \multicolumn{1}{c|}{0.5\% (0.18\%)} \\ \hline \hline
\multirow{3}{*}{\textbf{T = 32}} & \multicolumn{1}{c|}{0.10}   & \multicolumn{1}{c||}{88.9\% (1.60\%)}   & \multicolumn{1}{c|}{20.9\% (4.98\%)}          &                                  \multicolumn{1}{c|}{7.8\% (2.60\%)}   \\ \cline{2-5} 
                                & \multicolumn{1}{c|}{0.25}   & \multicolumn{1}{c||}{75.2\% (1.50\%)}   & \multicolumn{1}{c|}{8.0\% (1.65\%)}          &                                  \multicolumn{1}{c|}{3.1\% (1.15\%)}  \\ \cline{2-5} 
                                & \multicolumn{1}{c|}{0.50}    & \multicolumn{1}{c||}{50.0\% (1.78\%)}    & \multicolumn{1}{c|}{1.8\% (0.75\%)}           &                                   \multicolumn{1}{c|}{0.5\% (0.45\%)} \\ \hline
\end{tabular}
\caption{{\small\textbf{Evaluating human judgment of human behavior.}
\ian{Using data provided by~\cite{devlin2021navigation}, we evaluate the human-like navigation behavior of various agents.
We run each experiment 10 times and report the median and IQR of the $p$-value.
Note that the human-to-human likeness experiments are evaluated using random splits of the human player data, while the human-to-hybrid and human-to-symbolic evaluations use the full distributions of movements collected from each agent.}}}
\label{tbl:human_judgement}
\end{table}

In their study of human-like navigation behavior, Devlin \textit{et al.}~\cite{devlin2021navigation} trained two reinforcement learning-based (RL-based) agents, which they refer to as \textit{symbolic} and \textit{hybrid} agents\footnote{Please refer to~\cite{devlin2021navigation} for further details on the architectures, observation spaces, and training processes for these agents.}, to complete a navigation task in a modern AAA video game.
The task is set up as a 3D game map with 16 possible target locations.
For each episode, the agent must navigate to a target that is spawned uniformly at random in one of the 16 locations.
Devlin \textit{et al.}~\cite{devlin2021navigation} trained the two agents to achieve a level of task success sufficiently similar to that of human players so as to focus solely on the learned behavior.
To analyze human judgment of human-like behavior, they design a human Navigational Turing Test (HNNT) in which they administer a survey to 60 human assessors.
They report not only that participants were able to accurately detect human players with statistical significance, but also that the behavior of the hybrid agent was judged to be more human-like when directly compared to the behavior of the symbolic agent.
In fact, participants judged the hybrid agent to be more human-like 78\% of the time on average.
For the purpose of evaluating the efficacy of our hypothesis test when used to analyze the human-like behavior of artificial agents, we benchmark our results using the data reported in their study.

Devlin \textit{et al.}~\cite{devlin2021navigation} provide a dataset composed of 100 episodes anonymously collected from a pool of human players and 50 episodes separately collected from pre-trained symbolic and hybrid agents rolled out at two unique checkpoints.
Following the process detailed in Section~\ref{sec:test}, we generate our sample distribution of human trajectories by independently bootstrapping $K$ trajectories of length $T$ from each of the 100 episodes collected from the human players.
Similarly, we independently generate our sample distribution of symbolic and hybrid agent trajectories from each of the 50 episodes collected from the latest checkpoints provided for each agent.
In our experiments, we observe that the longest episodes collected from the human players, hybrid agent, and symbolic agent were 1.9k, 1.1k, and 1.6k, respectively.

With our sample distributions, we first analyze our test's ability to measure human-to-human likeness.
Because the 100 episodes collected from the human players are anonymously pooled, we randomly split the set of episodes into two unique sets when comparing human-to-human likeness.
In each test, we use different random splits of the data and use $S=1000$ iterations with a subsample size of $m=250$.
We repeat each experiment 10 times and report the median $p$-value and interquartile range (IQR) across different time horizon intervals ($T$) and alpha levels ($\alpha$) in the first column of Table~\ref{tbl:human_judgement}.
We observe that the $p$-value is nearly $1 - \alpha$ across all selections of $T$ and $\alpha$ when comparing two unique splits of the data collected from human players.

Finally, we apply our hypothesis test to compare the behaviors of the symbolic and hybrid agents to those of human players.
For each experiment, we again use $S=1000$ with a subsample size of $m=250$.
Similarly, we repeat each experiment 10 times and report the median $p$-value and IQR across different time horizon intervals ($T$) and alpha levels ($\alpha$) in Table~\ref{tbl:human_judgement}.
Unlike the previous test which compares random splits of human data against each other, we use the full 100 episodes collected from humans to compare against the 50 episodes collected from each agent.
We observe that our test aligns with the human judgment reported in~\cite{devlin2021navigation} as the hybrid agent is determined to be more human-like than the symbolic agent across selections of $T$ and $\alpha$.
Note that across all selections of $T$ and $\alpha$, the human-to-human likeness is greater than the human-likeness of both the hybrid and symbolic agents.

\section{Discussion}
\label{sec:discussion}

\ian{The study of behavior, particularly human-like behavior, has a rich history that intersects across a variety of domains (\textit{e.g.}, robotics, psychology, gaming, \textit{etc.}), each with a unique vantage point of the same larger problem.
Yet, understanding the driving factors for emulating human-likeness still remains an open challenge.
While we focus this study on human-like navigation behavior in the context of gaming, the evaluation of behavior as a distribution of patterns may be extended across domains, abilities, and tasks.
Furthermore, while we show that our testing framework can effectively discern nuanced differences in navigation behavior, we do not investigate what differentiates such behaviors.}
However, we observe that our results seem to be robust to our selections of time horizon ($T$), which could suggest either that the evaluation of navigation behavior is not overly sensitive to variations in time horizons or that further exploration is needed into significantly larger values.
Nevertheless, as humans tend to be ``creatures of habit", we conjecture that analyzing distributions of behavior through the lens of fixed-length time horizons could yield further insights into the driving factors of human-like behavior.
We leave this exploration as well as extensions into domains such as reward design or imitation learning for future work.

\section{Conclusion and Future Work}
\label{sec:conclusion}

{We proposed a non-parametric two-sample hypothesis testing framework based on statistical resampling methods for the purpose of comparing the navigation behaviors of artificial agents to those of human players.
We showed that the $p$-value resulting from our test can be used as a measure of similarity (Section~\ref{sec:method}).
Furthermore, we demonstrate that our test is agnostic to the nature of the agent by comparing the human-like navigation behavior of RL-based and NavMesh-based agents in a custom environment (Section~\ref{sec:experiments}).
Finally, we use the human Navigational Turing Test (HNNT) results reported by~\cite{devlin2021navigation} to show that this rank ordering aligns with human judgment (Section~\ref{sec:human_judgement}).
While we focus on navigation, we believe that both our method and our central hypothesis have potential impact across other aspects of human-like behavior.

\pagebreak

\section*{Acknowledgements}
We would like to thank Gabor Sines, Thomas Perry, Alex Cann, Tian Yue Liu, and the rest of the AMD Software Technology and Architecture team for insightful discussions and infrastructure support. \\

\noindent © 2022 Advanced Micro Devices, Inc.  All rights reserved.
AMD, the AMD Arrow logo, Radeon, and combinations thereof are trademarks of Advanced Micro Devices, Inc.
Other product names used in this publication are for identification purposes only and may be trademarks of their respective companies.

\bibliography{citations.bib}

\end{document}